\documentclass[runningheads]{llncs}

\usepackage{mathrsfs}
\usepackage{array}
\usepackage[T1]{fontenc}
\usepackage{graphicx}
\usepackage{verbatim}
\usepackage{multirow}
\usepackage{hyperref}
\usepackage{amssymb}
\usepackage{amsmath}
\usepackage{pifont}
\newcommand{\cmark}{\ding{51}}

\usepackage{subcaption}
\usepackage{makecell}
\usepackage{marvosym}

\begin{document}

\title{Towards Holistic Surgical Scene Graph}

\author{
Jongmin Shin$^{\dag}$\inst{1}\and
Enki Cho$^{\dag}$\inst{2}\and
Ka Young Kim$^{\dag}$\inst{2}\and \\
Jung Yong Kim\inst{1}\and
Seong Tae Kim\textsuperscript{${\ast}$}\inst{2}
$^{\href{mailto:st.kim@khu.ac.kr}{\textrm{\Letter}}}$
\and Namkee Oh\textsuperscript{${\ast}$}\inst{1}
$^{\href{mailto:namkee.oh@samsung.com}{\textrm{\Letter}}}$
}

\authorrunning{Shin, Cho, Kim et al.}

\institute{
Department of Surgery, Samsung Medical Center, Seoul 06351, Republic of Korea \and
Kyung Hee University, Yongin 17104, Republic of Korea
}



\def\thefootnote{$\dag$}\footnotetext{Equal contribution; \textsuperscript{${\ast}$}~Corresponding author.}

\maketitle              

\begin{abstract}
Surgical scene understanding is crucial for computer-assisted intervention systems, requiring visual comprehension of surgical scenes that involves diverse elements such as surgical tools, anatomical structures, and their interactions. To effectively represent the complex information in surgical scenes, graph-based approaches have been explored to structurally model surgical entities and their relationships.
Previous surgical scene graph studies have demonstrated the feasibility of representing surgical scenes using graphs. 
However, certain aspects of surgical scenes—such as diverse combinations of tool-action-target and the identity of the hand operating the tool—remain underexplored in graph-based representations, despite their importance.
To incorporate these aspects into graph representations, we propose Endoscapes-SG201 dataset, which includes annotations for tool–action– target combinations and hand identity.
We also introduce SSG-Com, a graph-based method designed to learn and represent these critical elements.
Through experiments on downstream tasks such as critical view of safety assessment and action triplet recognition, we demonstrated the importance of integrating these essential scene graph components, highlighting their significant contribution to surgical scene understanding. The code and dataset are available at \url{https://github.com/ailab-kyunghee/SSG-Com}.


\end{abstract}

\section{Introduction}
\label{sec:intro}

\begin{table}[t]
\centering
\caption{Comparison of information types contained in graphs generated from each dataset.   Endoscapes-SG201 is our proposed dataset, which is based on Endoscapes-Bbox201~\cite{murali2023endoscapes} with additional annotations and refined labels.   
The numbers in parentheses (\#) denote the number of classes in each category.}
\label{tab:dataset_comparison}
\scalebox{0.9}{
\begin{tabular}{c|c|ccccc} 
\hline
\multirow{3}{*}{\textbf{Methods}} & \multirow{3}{*}{\textbf{Training Dataset}} & \multicolumn{5}{c}{\textbf{Contained Information}} \\ \cline{3-7}
& & \textbf{Surgical} & \textbf{Spatial} & \multirow{2}{*}{\textbf{Anatomy}} & \multirow{2}{*}{\textbf{Action}} & \textbf{Hand} \\
& & \textbf{Tool} & \textbf{Relation} &  &  & \textbf{ID} \\\hline
Islam et al.~\cite{islam2020learning} & EndoVis-18~\cite{allan20202018roboticscenesegmentation}    & \cmark(9)  & -         & \cmark(1) & \cmark(12) & -         \\
Holm et al.~\cite{holm2023dynamic}    & CaDIS~\cite{grammatikopoulou2022cadiscataractdatasetimage} & \cmark(12) & -         & \cmark(4) & -          & \cmark(1)        \\
LG-CVS~\cite{murali2023latent} & Endoscapes-Bbox201~\cite{murali2023endoscapes}                     & \cmark(1)  & \cmark(3) & \cmark(5) & -          & -         \\
SSG-Com (Ours)                & Endoscapes-SG201~(Ours)                     & \cmark(6)  & \cmark(3) & \cmark(5) & \cmark(6)  & \cmark(3) \\ \hline
\end{tabular}
}
\end{table}
Surgical scene understanding is crucial for computer-assisted surgery, as it captures diverse visual information—including surgical tools, anatomical structures, and their spatial and functional relationships within surgical scenes. The information captured from surgical scenes plays a key role in various surgical applications, including workflow analysis~\cite{Murali_2023,sharma2023surgical,valderrama2022towards}, automated critical view of safety (CVS) assessment~\cite{murali2023latent}, and surgical report generation~\cite{lin2023sgt++}.

To represent the complex information within surgical scenes, graph-based approaches~\cite{holm2023dynamic,islam2020learning,murali2023latent} have been explored to structurally model surgical entities and their relationships.
Islam \textit{et al.}~\cite{islam2020learning} generate scene graphs representing a single anatomical structure, surgical tools, and their interactions by using  EndoVis-18~\cite{allan20202018roboticscenesegmentation} dataset enhanced with bounding boxes and interaction annotations. 
Holm \textit{et al.}~\cite{holm2023dynamic} generated scene graphs that include fine-grained surgical tools and anatomy~\cite{grammatikopoulou2022cadiscataractdatasetimage}. 
In~\cite{murali2023latent}, Murali \textit{et al.} utilized the Endoscapes-Bbox201~\cite{murali2023endoscapes} dataset, which is annotated with bounding boxes for fine-grained anatomical structures. 

While previous studies have demonstrated the feasibility of representing surgical scene information using graphs, certain aspects of surgical scenes remain underexplored in graph representations. One of the aspects is the intricate interactions among various surgical tools and anatomical structures within the surgical scene. Given that a surgical scene inherently comprises diverse combinations of tools, actions, and targets~\cite{nwoye2022rendezvous}, it is possible to represent the relationships among tool-action-target pairs as a graph. Another important aspect is the identification of the hands operating the tools. Prior studies have primarily designated surgical tools as the nodes in graphs without explicitly reflecting the roles of medical personnel, such as operators and assistants. However, since each operating hand performs specific actions on the target, such as the surgeon's right hand executing a critical procedure while the assistant's hand retracts surrounding tissue, the hand identity constitutes essential information that should be explicitly incorporated into the scene graph.

To incorporate these underexplored yet crucial aspects of surgical scenes into graph representations, we propose a novel dataset for surgical scene graph, named Endoscapes-SG201. Figure~\ref{fig:SG201_Construction} illustrates the dataset construction process. 
Specifically, we refined the bounding boxes, subdivided the ``tool'' class into 6 classes—Grasper, Hook, Clipper, Bipolar, Irrigator, and Scissors—and annotated action and hand identity labels. 
The entire annotation process was iteratively reviewed to ensure accuracy and consistency, and the annotation of Endoscapes-SG201 will be publicly released to facilitate further research. 
Through experiments on two downstream tasks such as CVS assessment and Action Triplet Recognition, we demonstrated the significance of incorporating these previously unexplored scene graph components, highlighting their relevance to surgical scene understanding.
Our main contributions are as follows:
\begin{itemize}
    \item We introduce a novel surgical scene graph dataset, called Endoscapes-SG201. Based on Endoscapes-Bbox201, we refined bounding boxes, subdivided the `tool’ class into six classes, and annotated tool-action-target and hand identity labels.

    \item We propose SSG-Com, a graph-based method that explicitly integrates not only surgical entities and their spatial relationships but also action relationships and hand identity.
    
    \item We demonstrate the effectiveness of incorporating previously underexplored scene graph components—including tool–action–target combinations and hand identity—through evaluations on two downstream tasks such as CVS assessment and action triplet recognition.

\end{itemize}

\section{Endoscapes-SG201 Dataset}
\label{sec:Endoscapes-SG201 Dataset}

\begin{figure*}[t]
    \centering
    \includegraphics[width=\textwidth]{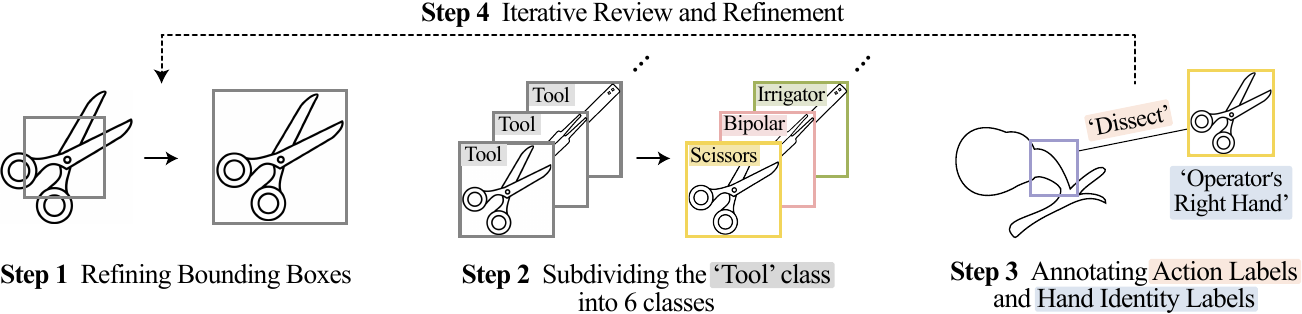}
    \caption{Overview of the Endoscapes-SG201 construction process. Based on Endoscapes-Bbox201, we refined box labels, subdivided the `tool' class into 6 classes, and annotated action and hand identity labels. Iterative review and refinement were conducted to enhance the dataset quality.}
    \label{fig:SG201_Construction} 
\end{figure*}
\sloppy
In this section, we describe the composition and construction process of Endoscapes-SG201, a novel dataset designed for comprehensive surgical scene graph generation. Endoscapes-SG201 is built upon Endoscapes-BBox201~\cite{murali2023endoscapes}, which consists of 1,933 frames extracted from 201 laparoscopic cholecystectomy videos. As in Table~\ref{tab:dataset_comparison}, Endoscapes-SG201 provides annotations for 6 surgical tools, 5 anatomical structures, 6 actions, and 3 hand identities. Bounding box annotations are provided for both surgical tools and anatomical structures, while action annotations are provided in the form of action triplets (tool, action, target). In addition, hand identity annotations are provided for each surgical tool. 

To construct annotations based on Endoscapes-BBox201, we followed the structured procedure illustrated in Figure~\ref{fig:SG201_Construction}. First, we refined the bounding boxes of surgical tools that were either missing or inaccurately labeled in Endoscapes-BBox201, improving the precision of tool localization. Second, while anatomical structures were annotated with 5 classes, all surgical tools were grouped under a single generic class, ``tool,'' which made it difficult to distinguish functional differences between tools. To address this limitation, we subdivided the ``tool'' class into 6 classes—Hook, Grasper, Clipper, Bipolar, Irrigator, and Scissors. Third, we added hand identity annotations to indicate which hand is operating each surgical tool. Each tool was assigned to one of the following categories: surgeon’s right hand (Rt), surgeon’s left hand (Lt), or assistant’s hand (Assi). All annotation processes were conducted by two experts, ensuring annotation consistency through iterative review and correction. Following the dataset partitioning scheme used in~\cite{murali2023endoscapes}, we divided Endoscapes-SG201 into 1,212 training frames, 409 validation frames, and 312 test frames. The detailed distribution of the Endoscapes-SG201 dataset is presented in Table~\ref{tab:category_comparison}.
\sloppy

\begin{table}[t]
\caption{Category-wise distribution of surgical tools, actions, and manipulating hands in the dataset. Surgical tools consist of Hook (HK), Grasper (GP), Clipper (CL), Bipolar (BP), Irrigator (IG), and Scissors (SC). Actions consist of Dissect (Dis.), Retract (Ret.), Grasp (Gr.), Clip (Cl.), Coagulate (Co.), and Null\_verb (Null). Hand Identity consist of operator's right hand (Rt), left hand (Lt), and the assistant's hand (Assi).}
\centering
\scalebox{0.95}{
\begin{tabular}{c|cccccc|cccccc|ccc} 
\hline
\multirow{2}{*}{\textbf{Set}} & \multicolumn{6}{c|}{\textbf{Surgical Instruments}} & \multicolumn{6}{c|}{\textbf{Action}} & \multicolumn{3}{c}{\textbf{Hand Identity}} \\ \cline{2-16}
& \textbf{HK} & \textbf{GP} & \textbf{CL} & \textbf{BP} & \textbf{IG} & \textbf{SC} & \textbf{Dis.} & \textbf{Ret.} & \textbf{Gr.} & \textbf{Cl.} & \textbf{Co.} & \textbf{Null.} & \textbf{Rt} & \textbf{Lt} & \textbf{Assi} \\ \hline

\textbf{Train}  & 686 & 997 & 128 & 95 & 41 & 3 & 601 & 879 & 72 & 122 & 41 & 233 & 986 & 842 & 122 \\
\textbf{Val}    & 202 & 347 & 48  & 36 & 11 & 0 & 168 & 308 & 12 & 44 & 14 & 98 & 311 & 268 & 65 \\
\textbf{Test}   & 172 & 254 & 43  & 11 & 17 & 1 & 147 & 242 & 11 & 41 & 5 & 52 & 246 & 218 & 34 \\ \hline
\textbf{Total}  & 1060 & 1598 & 219 & 142 & 69 & 4 & 916 & 1429 & 95 & 207 & 60 & 383 & 1543 & 1328 & 221 \\
\hline
\end{tabular}}
\label{tab:category_comparison}
\end{table}

\section{Methodology}
\label{sec:method}

\begin{figure*}[t]
    \centering
    \includegraphics[width=\textwidth]{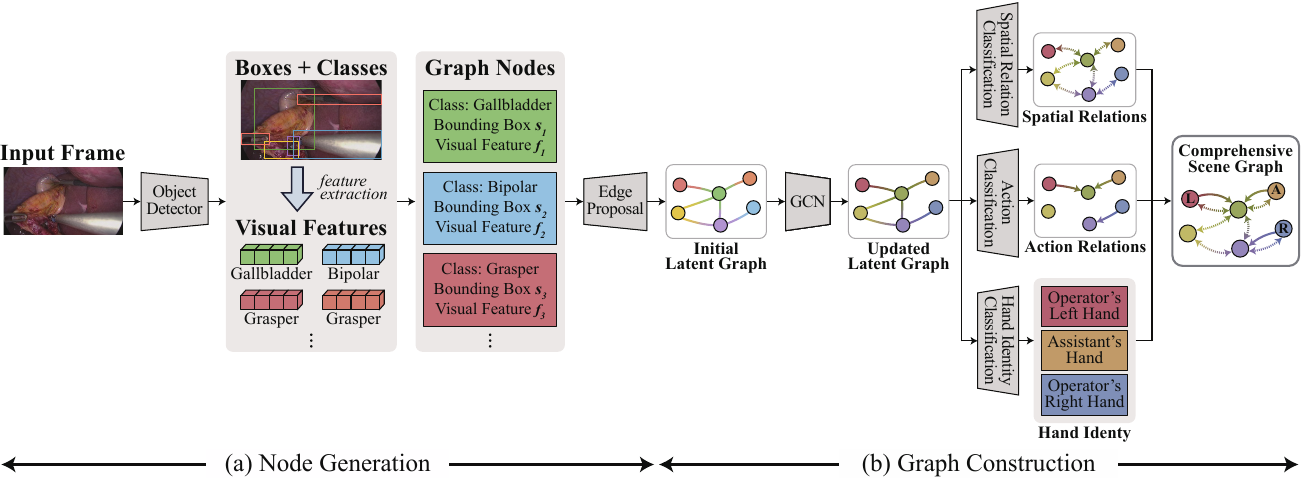}
    \caption{Overview of the proposed surgical scene graph generation.}
    \label{fig:SG_main_fig} 
\end{figure*}

In this study, we propose SSG-Com (Surgical Scene Graph for Comprehensive Understanding), a graph-based method that explicitly considers not only the relationships between surgical tools and anatomical structures but also incorporates information about the hand identity operating each tool. 
We adopt LG-CVS~\cite{murali2023latent} as our baseline model, which establishes edges primarily based on spatial relationships (left-right, above-below, inside-outside) between detected objects. 
In contrast, our proposed method extends this approach by introducing additional edges called Surgical Action Edges (SAE), explicitly capturing interactions between surgical tools and anatomical structures, thus enabling a more contextually informed understanding of relationships. In the first stage, as shown in Figure~\ref{fig:SG_main_fig}, an object detection model identifies surgical tools and anatomical structures, which are then set as graph nodes. The edge feature between two nodes is constructed based on the union region of the two corresponding objects. Through edge proposal, only the meaningful edges are retained, forming the initial latent graph. This graph is then passed through a Graph Convolutional Network (GCN), resulting in an updated latent graph. 
In the second stage, the updated latent graph is fed to task-specific classifiers for downstream tasks (e.g., CVS prediction or triplet recognition).

\subsection{Graph Construction}
The graph \( \mathcal{G} = (\mathcal{N}, \mathcal{E}) \) is constructed by defining detected objects as nodes \( \mathcal{N} \) and establishing edges \( \mathcal{E} \) that represent their relationships.
Each node is characterized by its bounding box coordinates \( p \), class probability \( s \), and feature vector \( f \), and categorized as either a surgical tool node or an anatomy node as follows:
\begin{equation}
    \mathcal{N}_{\text{tool}} = \{ (p_i, s_i, f_i) \mid s_i \in \mathcal{C}_{\text{tool}} \}, \quad
    \mathcal{N}_{\text{anatomy}} = \{ (p_j, s_j, f_j) \mid s_j \in \mathcal{C}_{\text{anatomy}} \},
\end{equation}
where \( f_i \) is the feature vector extracted via RoIAlign~\cite{he2017mask}, and \( \mathcal{C}_{\text{tool}} \), \( \mathcal{C}_{\text{anatomy}} \) are the class sets for surgical tools and anatomical structures, respectively. LG-CVS~\cite{murali2023latent} proposed a method for modeling spatial relationships using node position-based edges (e.g., left-right, above-below, inside-outside). 
While effective for capturing spatial layouts, this approach lacks contextual understanding of surgical interactions. To address this, we introduce Surgical Action Edges (SAE), which explicitly encode meaningful interactions between surgical tools and anatomical structures. An edge \textit{e} is formed as follows:
\begin{equation}
    e_{i,j}^{\text{action}} =
    \begin{cases} 
        1, & \text{if } s_i \in \mathcal{C}_{\text{tool}} \text{ and } s_j \in \mathcal{C}_{\text{anatomy}} \\
        0, & \text{otherwise}.
    \end{cases}
\end{equation}
This graph structure enables the model to move beyond spatial relationships, incorporating explicit representations of surgical actions.

\subsection{Surgical Action Edge Classification}
To learn action relationships, we introduce an independent surgical action edge classifier to explicitly capture how surgical tools interact with anatomical structures.
For each edge \( e_{i,j} \), the edge feature vector \( f_{i,j} \) is passed through the action edge classifier \( \mathcal{C}_{\text{action}}(\cdot) \) to predict the action type, resulting in \( \hat{y}_{i,j}^{\text{action}} = \mathcal{C}_{\text{action}}(f_{i,j}) \). 
The classifier is formulated as a multi-class classification problem with 6 action classes, as in Table~\ref{tab:category_comparison}. 
The loss function for training the classifier is defined using the cross-entropy loss as \( \mathcal{L}_{\text{action}} = \text{CE}(y_{i,j}^{\text{action}}, \hat{y}_{i,j}^{\text{action}}) \).
This classification enables the model to recognize how surgical tools interact with anatomical structures, leading to a more refined understanding of surgical procedures.

\subsection{Hand Identity Classification}
To better capture contextual information on surgical tool usage, SSG-Com consider not only the relationships between surgical tools and anatomical structures but also the hand operating the tool (left hand, right hand, or assistant hand). To achieve this, we introduce a classifier that predicts the hand identity using the feature of the tool node \( \mathcal{N}_{\text{tool}} \). 
This process is defined as \( \hat{y}_{i}^{\text{hand}} = \mathcal{C}_{\text{hand}}(f_i) \), where \( \mathcal{C}_{\text{hand}}(\cdot) \) is a classifier designed to predict the hand identity class. The model is trained using the cross-entropy loss \( \mathcal{L}_{\text{hand}} = \text{CE}(y_{i}^{\text{hand}}, \hat{y}_{i}^{\text{hand}}) \).
By incorporating hand identity classification, the model learns not only the operation of surgical tools but also which hand is used to operate each tool, enhancing the overall contextual understanding of surgical scenes.

The total training objective is defined by combining the latent graph learning loss \( \mathcal{L}_{LG} \) with surgical action edges classification loss \( \mathcal{L}_{action} \) and hand identity classification loss \( \mathcal{L}_{hand} \). Here, \( \mathcal{L}_{LG} \) refers to the loss function proposed in~\cite{murali2023latent}, which consists of three components: the loss function of the object detection model, the binary cross-entropy loss to determine the existence of meaningful edges between nodes, and the cross-entropy loss to classify spatial relation. Finally, the total loss function used for training, \( \mathcal{L}_{total} \), is defined as follows:
\begin{equation} 
    \mathcal{L}_{\text{total}} = \mathcal{L}_{LG} + \lambda_{action} \cdot \mathcal{L}_{action} + \lambda_{hand} \cdot \mathcal{L}_{hand},
\end{equation}
where \( \lambda_{action} \) and \( \lambda_{hand} \) are hyperparameters that balance the importance of surgical action edge learning and hand identity classification, respectively.

\subsection{Graph-Based Learning for Downstream Tasks}
For downstream tasks, the object detector within the pretrained latent graph encoder from the first stage is frozen, and the remaining components are fine-tuned. Depending on the task (e.g., CVS prediction or Triplet Recognition), the corresponding decoder is designed and trained using task-specific labels.

\section{Experiments}

\subsection{Experimental Setting}
\noindent \textbf{CVS Prediction.}
\label{subsubsec:cvs_pred}
The CVS consists of three independent criteria: C1.Two Structures, C2.HCT Dissection, and C3.Cystic Plate, making it a multi-label classification task. For our experiments, we use the Endoscapes-CVS201~\cite{murali2023endoscapes}, which contains 11,090 images uniformly sampled from the dissection phase of 201 cholecystectomy procedures. The dataset is split into 6,960 training images, 2,331 validation images, and 1,799 test images. For evaluation, we report mean average precision (mAP) across the three CVS criteria for all methods.

\noindent \textbf{Triplet Recognition.}
\label{subsubsec:triplet_recog}
Triplet recognition aims to classify 34 distinct triplet labels that can appear in any annotated frame within the Endoscapes-SG201. For this experiment, we use 1,933 images that include box annotations, ensuring precise localization of relevant objects for triplet recognition. The dataset is divided into 1,212 training images, 409 validation images, and 312 test images. For evaluation, we measure mAP across all 34 triplet labels, which represent possible cases that can appear in the surgical video frames.

\noindent \textbf{Implementation Details.} 
All experiments were conducted on a single NVIDIA RTX 3090 GPU. We used Faster R-CNN~\cite{ren2015faster} as the object detector. Each training stage lasted 50 epochs, and the best weights were selected based on the validation performance. \( \lambda_{action}\) and \( \lambda_{hand}\) are set to 0.6 and 0.001, respectively.

\begin{table*}[t]
\centering
\label{tab:evaluation_results}
\caption{Comparison of methods in triplet recognition and CVS prediction. Spa., Act., and Hand denote the types of information incorporated in the graph: spatial relationship, action relationship, and hand identity, respectively.}
\vspace{-5pt}
\begin{minipage}{0.59\linewidth}
    \centering
    \subcaption{Triplet Recognition Performance}
    \label{tab:ablation_triplet}
    \scalebox{0.85}{
    \begin{tabular}{c|c|ccc|c}
    \hline
     \multirow{2}{*}{Dataset}&\multirow{2}{*}{Method} & \multicolumn{3}{c|}{Graph} & Triplet \\ 
    \cline{3-5}
     & & Spa. & Act. & Hand & (mAP) \\ \hline
    \multirow{2}{*}{\makecell{{Endoscape} \\ {-BBox201~\cite{murali2023endoscapes}}}}
    & ResNet50-DetInit   &  &  &   & 9.2   \\
    & LG-CVS             & \cmark &  & & 13.8  \\
    \hline
    \multirow{4}{*}{\makecell{{Endoscape} \\ {-SG201}}}
    & ResNet50-DetInit   &  &  &   & 9.7   \\
    & LG-CVS             & \cmark &  & & 18.0  \\
    & SSG-Com (Ours)      & \cmark & \cmark & & 23.5\\
    & SSG-Com (Ours)      & \cmark & \cmark & \cmark & \textbf{24.2}\\ \hline
    \end{tabular}
    }
\end{minipage}
\hfill
\begin{minipage}{0.4\linewidth}
    \centering
    \subcaption{CVS Performance}
    \label{tab:ablation_CVS}
    \scalebox{0.8}{
    \begin{tabular}{c|c|c}
    \hline
    \multirow{2}{*}{Dataset} & \multirow{2}{*}{Method} & CVS\\  
    & & (mAP)\\ \hline
    \multirow{4}{*}{\makecell{{Endoscape} \\ {-BBox201~\cite{murali2023endoscapes}}}}
    & DeepCVS                     & 42.4 \\
    & LayoutCVS                   & 42.8 \\
    & ResNet50-DetInit            & 57.3 \\
    & LG-CVS                      & 62.3 \\
    \hline
    \multirow{5}{*}{\makecell{{Endoscape} \\ {-SG201}}}
    & DeepCVS                     & 40.7 \\
    & LayoutCVS                   & 44.0 \\
    & ResNet50-DetInit            & 55.3 \\
    & LG-CVS                      & 63.2 \\
    & SSG-Com (Ours)               & \textbf{64.6} \\ \hline
    \end{tabular}
    }
\end{minipage}
\end{table*}

\subsection{Comparison with Other Methods}
\begin{figure*}[t]
    \centering
    \includegraphics[width=0.95\textwidth]{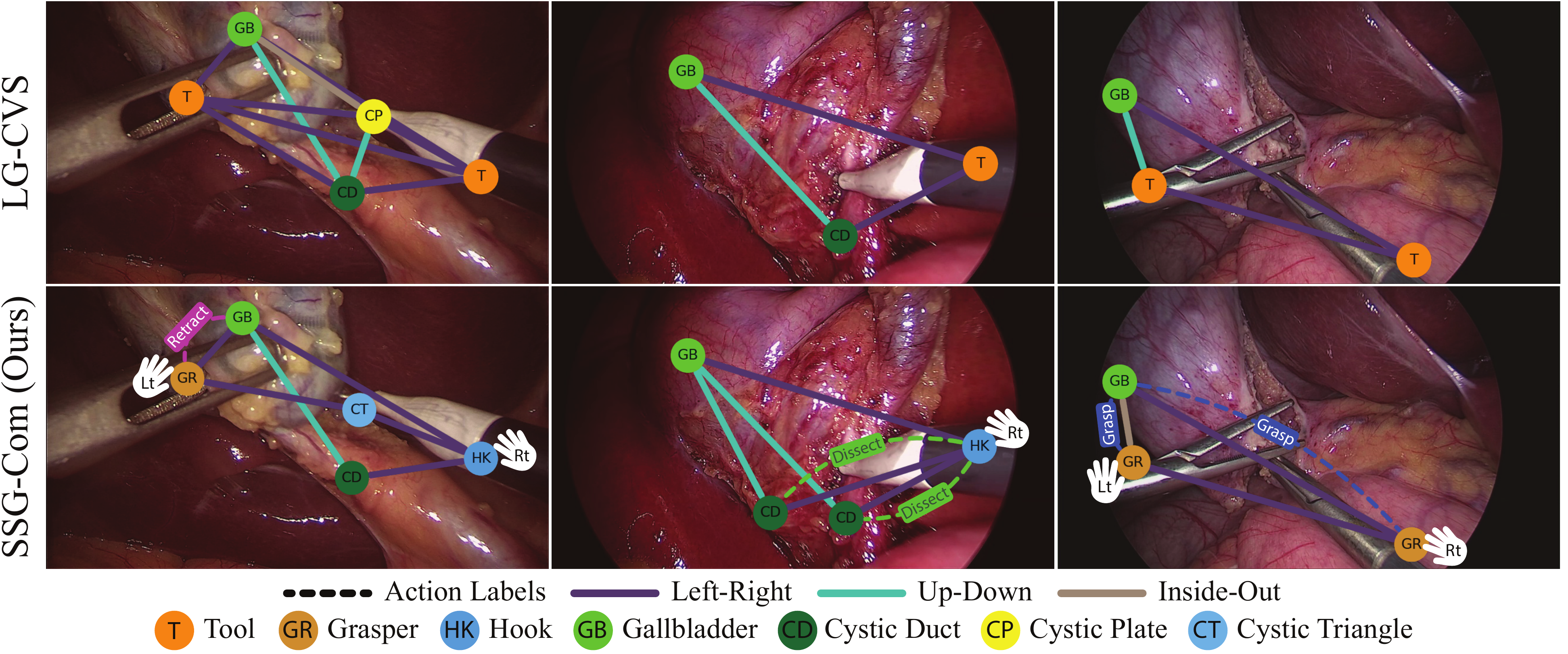}
    \caption{Qualitative results of the graphs generated by LG-CVS and SSG-Com. Nodes represent anatomical structures and surgical tools. Edges depict relationships: solid straight lines indicate spatial relationships, while dashed curved lines indicate action relationships. A hand next to tool nodes denotes the hand identity operating the tool.}
    \label{fig:SG_quali_fig} 
\end{figure*}

In this section, we compare the performance of the proposed method with other methods in triplet recognition and CVS prediction. For triplet recognition, we compare our method with LG-CVS~\cite{murali2023latent} and ResNet50-DetInit. LG-CVS learns a latent graph representation incorporating spatial relations from object detection results. ResNet50-DetInit~\cite{murali2023latent}, is a multi-task learning approach. As shown in Table~\ref{tab:ablation_triplet}, SSG-Com achieves 24.2 mAP, significantly outperforming LG-CVS (18.0 mAP) and ResNet50-DetInit (9.7 mAP). For CVS prediction, we further compare against DeepCVS~\cite{mascagni2022artificial} and LayoutCVS. LayoutCVS, introduced in~\cite{murali2023latent}, follows the DeepCVS architecture but uses only layout information as input, excluding the original image. All models undergo a two-stage training process, first on SG201 and then on Endoscapes-CVS201~\cite{murali2023endoscapes} for fine-tuning and evaluation. As shown in Table~\ref{tab:ablation_CVS}, SSG-Com achieves 64.6 mAP, surpassing LG-CVS (63.2 mAP) and ResNet50-DetInit (55.3 mAP). 
Figure~\ref{fig:SG_quali_fig} shows qualitative results of LG-CVS and SSG-Com.
These results demonstrate that our graph-based representation models surgical scenes more effectively, leading to superior performance in both Triplet Recognition and CVS prediction compared to other methods.

\subsection{Ablation study}
We further analyze the impact of model learning strategies and dataset composition on performance, evaluating the significance of incorporating surgical-related information. Table~\ref{tab:ablation_triplet} compares model performance based on the presence of graph-based learning and the diversity of information incorporated into the graph. ResNet50-DetInit utilizes only visual information, LG-CVS~\cite{murali2023latent} learns a latent graph considering only spatial relationships, while the proposed method incorporates both surgical actions and operating hand information. As shown in the table, models trained with graph representations outperform those relying solely on visual features, and increasing the diversity of information within the graph further improves performance. Furthermore, an ablation study within the proposed method reveals that including the subject performing the action (operating hand) improves performance compared to using only action information. 
%
%
Table~\ref{tab:ablation_triplet} and Table~\ref{tab:ablation_CVS} compare the performance of models trained on the Endoscapes-BBox201~\cite{murali2023endoscapes} and Endoscapes-SG201. The results show that models trained on Endoscapes-SG201 consistently achieve higher performance across downstream tasks, confirming that the proposed dataset provides more refined and informative training data. In the Triplet recognition task, training with individual surgical tool classes yields superior performance compared to treating all tools as a single class. This suggests that fine-grained tool classification enhances model learning and enables more precise relationship modeling.

\section{Conclusion}
\label{sec:conclusions}
In this study, we introduced a novel Endoscapes-SG201 dataset to build a holistic surgical scene graph. 
To represent the complex information within surgical scenes, our method incorporates various relations, including action and spatial relations between tool and target anatomy. We also include hand identity to represent the surgical scenes. 
Comparative experiments have been conducted on two downstream tasks of CVS assessment and action triplet recognition. Experimental results show that our method is effective in representing complex surgical scenes, which leads to achieving higher performance compared with previous studies. We believe that the Endoscapes-SG201 dataset has the potential to contribute to future research in the computer-assisted intervention community by advancing holistic surgical scene understanding.

\subsubsection{Acknowledgements.} 
This work was supported by the National Research Foundation of Korea (NRF) grants funded by the Korea government (MSIT) (No. RS-2024-00334321, RS-2024-00392495), the Institute of Information \& Communications Technology Planning and Evaluation (IITP) grant funded by the Korea government (MSIT) (No. RS-2024-00509257, Global AI Frontier Lab), and the `Future Medicine 2030 Project' of Samsung Medical Center (No. SMX1230771).

\subsubsection{\discintname}
The authors have no competing interests to declare that are relevant to the content of this article. 

%
%
%
\bibliographystyle{splncs04}
\bibliography{main}

\begin{thebibliography}{10}
\providecommand{\url}[1]{\texttt{#1}}
\providecommand{\urlprefix}{URL }
\providecommand{\doi}[1]{https://doi.org/#1}

\bibitem{allan20202018roboticscenesegmentation}
Allan, M., Kondo, S., Bodenstedt, S., Leger, S., Kadkhodamohammadi, R., Luengo, I., Fuentes, F., Flouty, E., Mohammed, A., Pedersen, M., et~al.: 2018 robotic scene segmentation challenge. arXiv preprint arXiv:2001.11190  (2020)

\bibitem{grammatikopoulou2022cadiscataractdatasetimage}
Grammatikopoulou, M., Flouty, E., Kadkhodamohammadi, A., Quellec, G., Chow, A., Nehme, J., Luengo, I., Stoyanov, D.: Cadis: Cataract dataset for surgical rgb-image segmentation. Medical Image Analysis  \textbf{71},  102053 (2021)

\bibitem{he2017mask}
He, K., Gkioxari, G., Doll{\'a}r, P., Girshick, R.: Mask r-cnn. In: Proceedings of the IEEE international conference on computer vision. pp. 2961--2969 (2017)

\bibitem{holm2023dynamic}
Holm, F., Ghazaei, G., Czempiel, T., {\"O}zsoy, E., Saur, S., Navab, N.: Dynamic scene graph representation for surgical video. In: Proceedings of the IEEE/CVF international conference on computer vision. pp. 81--87 (2023)

\bibitem{islam2020learning}
Islam, M., Seenivasan, L., Ming, L.C., Ren, H.: Learning and reasoning with the graph structure representation in robotic surgery. In: Medical Image Computing and Computer Assisted Intervention--MICCAI 2020: 23rd International Conference, Lima, Peru, October 4--8, 2020, Proceedings, Part III 23. pp. 627--636. Springer (2020)

\bibitem{lin2023sgt++}
Lin, C., Zhu, Z., Zhao, Y., Zhang, Y., He, K., Zhao, Y.: Sgt++: Improved scene graph-guided transformer for surgical report generation. IEEE Transactions on Medical Imaging  (2023)

\bibitem{mascagni2022artificial}
Mascagni, P., Vardazaryan, A., Alapatt, D., Urade, T., Emre, T., Fiorillo, C., Pessaux, P., Mutter, D., Marescaux, J., Costamagna, G., et~al.: Artificial intelligence for surgical safety: automatic assessment of the critical view of safety in laparoscopic cholecystectomy using deep learning. Annals of surgery  \textbf{275}(5),  955--961 (2022)

\bibitem{murali2023endoscapes}
Murali, A., Alapatt, D., Mascagni, P., Vardazaryan, A., Garcia, A., Okamoto, N., Costamagna, G., Mutter, D., Marescaux, J., Dallemagne, B., et~al.: The endoscapes dataset for surgical scene segmentation, object detection, and critical view of safety assessment: official splits and benchmark. arXiv preprint arXiv:2312.12429  (2023)

\bibitem{Murali_2023}
Murali, A., Alapatt, D., Mascagni, P., Vardazaryan, A., Garcia, A., Okamoto, N., Mutter, D., Padoy, N.: Encoding surgical videos as latent spatiotemporal graphs for object and anatomy-driven reasoning. In: International Conference on Medical Image Computing and Computer-Assisted Intervention. pp. 647--657 (2023)

\bibitem{murali2023latent}
Murali, A., Alapatt, D., Mascagni, P., Vardazaryan, A., Garcia, A., Okamoto, N., Mutter, D., Padoy, N.: Latent graph representations for critical view of safety assessment. IEEE Transactions on Medical Imaging  (2023)

\bibitem{nwoye2022rendezvous}
Nwoye, C.I., Yu, T., Gonzalez, C., Seeliger, B., Mascagni, P., Mutter, D., Marescaux, J., Padoy, N.: Rendezvous: Attention mechanisms for the recognition of surgical action triplets in endoscopic videos. Medical Image Analysis  \textbf{78},  102433 (2022)

\bibitem{ren2015faster}
Ren, S., He, K., Girshick, R., Sun, J.: Faster r-cnn: Towards real-time object detection with region proposal networks. Advances in neural information processing systems  \textbf{28} (2015)

\bibitem{sharma2023surgical}
Sharma, S., Nwoye, C.I., Mutter, D., Padoy, N.: Surgical action triplet detection by mixed supervised learning of instrument-tissue interactions. In: International Conference on Medical Image Computing and Computer-Assisted Intervention. pp. 505--514. Springer (2023)

\bibitem{valderrama2022towards}
Valderrama, N., Ruiz~Puentes, P., Hern{\'a}ndez, I., Ayobi, N., Verlyck, M., Santander, J., Caicedo, J., Fern{\'a}ndez, N., Arbel{\'a}ez, P.: Towards holistic surgical scene understanding. In: International conference on medical image computing and computer-assisted intervention. pp. 442--452. Springer (2022)

\end{thebibliography}

\end{document}